%% file: 00_main.tex
\documentclass[runningheads]{llncs}
\usepackage[utf8]{inputenc}
\usepackage[english]{babel}

\usepackage{graphicx}
\usepackage{courier} 
\usepackage{url}
\usepackage{hyperref}
\usepackage{booktabs}
\usepackage{multirow}
\usepackage{cite}
\usepackage{makecell}
\usepackage{todonotes}
\usepackage{stmaryrd} 

\usepackage{fancyvrb}
\usepackage{cite}
\usepackage{algorithm,float}
\usepackage[noend]{algpseudocode}

\usepackage{listings}

\usepackage{algorithm,float}
\usepackage[caption=false]{subfig}
\usepackage[export]{adjustbox} 

\lstdefinelanguage{sparql} {
language={SQL},
    alsoletter={-},
    morekeywords={qkg,xsd,rdfs,skos,owl,dbr,rdf,so},
}

\newcommand{\kg}{\textit{Quote\-KG}}
\newcommand{\voc}[2]{\texttt{#1:\allowbreak #2}}
\newcommand{\schema}[1]{\texttt{#1}}

\begin{document}
\title{\kg{}: \\ A Multilingual Knowledge Graph of Quotes}
\titlerunning{\kg{}: A Multilingual Knowledge Graph of Quotes}
\author{Tin Kuculo\inst{1}\orcidID{0000-0001-6874-8881} \and Simon Gottschalk\inst{1}\orcidID{0000-0003-2576-4640} \and Elena Demidova\inst{2}\orcidID{0000-0002-5134-9072}}
\authorrunning{T. Kuculo et al.}
%
\institute{L3S Research Center, Leibniz Universität Hannover, Hannover, Germany \\
\email{\{kuculo,gottschalk\}@L3S.de}
\and
Data Science \& Intelligent Systems (DSIS), Universität Bonn, Bonn, Germany \\
\email{demidova@cs.uni-bonn.de}}

\maketitle              
\begin{abstract}

Quotes of public figures can mark turning points in history. A quote can explain its originator's actions, foreshadowing political or personal decisions and revealing character traits. Impactful quotes cross language barriers and influence the general population's reaction to specific stances, always facing the risk of being misattributed or taken out of context. The provision of a cross-lingual knowledge graph of quotes that establishes the authenticity of quotes and their contexts is of great importance to allow the exploration of the lives of important people as well as topics from the perspective of what was actually said.
In this paper, we present QuoteKG, the first multilingual knowledge graph of quotes. We propose the QuoteKG creation pipeline that extracts quotes from Wikiquote, a free and collaboratively created collection of quotes in many languages, and aligns different mentions of the same quote. 
QuoteKG includes nearly one million quotes in $55$ languages, said by more than $69,000$ people of public interest across a wide range of topics. QuoteKG is publicly available and can be accessed via a SPARQL endpoint.

\end{abstract}

\noindent
\textbf{Resource DOI: 10.5281/zenodo.4702544} \\
\textbf{Permanent URL:} \url{https://quotekg.l3s.uni-hannover.de} \\

\input{01_introduction}
\input{02_impact}
\input{03_schema}

\input{04_extraction}

\input{05_evaluation_statistics_examples}

\input{07_availability}
\input{08_related_work}
\input{09_conclusion}

\subsubsection*{Acknowledgements} 
This work was partially funded by H2020-MSCA-ITN-2018-812997 under ``Cleopatra''.

\newpage

\bibliographystyle{splncs04}
\bibliography{bibliography}

\end{document}

%% file: 01_introduction.tex
\section{Introduction}
\label{sec:introduction}

Quotes of public figures provide valuable information to understand their thoughts and attitudes, potentially leading to historically important actions, and thus serve as a crucial component in exploring world history~\cite{Thoughts}. Table~\ref{tab:intro_examples} provides three examples of quotes, with the first one emphasising the relevance of historic quotes: in 1930, Winston Churchill recognised the value of reading them. The second example in Table~\ref{tab:intro_examples} illustrates the relevance of quotes in world history: During a press conference in 2015, the German chancellor Angela Merkel said ``Wir schaffen das'' (``We can do this'') when the European migrant crisis unfolded and Germany prepared for the reception of refugees from Northern Africa and the Middle East. Since then, these three words defined Merkel's political course in the migrant crisis -- and led both to a welcoming culture as well as the rise of nationalist protests and right-wing political parties~\cite{mushaben2017wir,wirschaffendas}. 

Given this potential impact of words, it is of utmost importance to provide sources to quotes and to dismiss hoaxes~\cite{robinson2018did,keyes2007quote}: The third example in Table~\ref{tab:intro_examples} is a famous quote that has been attributed to different people, including Albert Einstein, Benjamin Franklin, and Mark Twain, but has not actually been said by any of them.\footnote{Reasons for false attribution of quotes to persons include to appear educated or to lend authority from the person.~\cite{reucher2021}.} In general, a quote can be mentioned in different sources, and mentions can deviate. For example, ``Wir schaffen das'' might be mentioned as ``We can do this'' or ``We will make it!'' in English translations. Therefore, there is a need to align mentions to the same quote and to provide context information such as the source and description (e.g., ``during a press conference'').

\setlength{\tabcolsep}{0.5em}
\renewcommand{\arraystretch}{1.5}
\begin{table}[th]
\caption{Three example quotes, together with their originators and dates. The last column gives examples of context that can be attributed to the mention of a quote, including source information, translations or validation of the quote's correctness.}
\label{tab:intro_examples}
\begin{tabular}{p{5.2cm}llp{3.2cm}} \toprule
\multicolumn{1}{c}{\textbf{Quote}} & \multicolumn{1}{c}{\textbf{By}} & \multicolumn{1}{c}{\textbf{Date}} & \multicolumn{1}{c}{\textbf{Selected Context}} \\ \midrule
It is a good thing for an uneducated man to read books of quotations. (en) & \makecell[tl]{Winston\\ Churchill} & 1930 & \textit{Source:} Roving Commission: My Early Life (1930) Chapter 9 \\ 
Wir schaffen das. (de) & \makecell[tl]{Angela \\ Merkel} & \makecell[tl]{2015, \\ Aug 31} & \textit{Translation:} We can do this (en) \\ 
The definition of insanity is doing the same thing over and over and expecting different results. (en) & \makecell[tl]{Albert\\ Einstein} &  & Misattributed. \\ \bottomrule
\end{tabular}
\end{table}

In this paper, we introduce \kg{} -- a new knowledge graph that provides nearly one million quotes said by more than $69,000$ persons of public interest in $55$ languages. Quotes in \kg{} come with detected sentiment and context such as their origin dates and sources. They are interlinked with their originators and other entities such as persons or events they refer to. Different mentions of the same quote are aligned across languages.

The creation of a knowledge graph covering quotes in many languages and their contexts faces several challenges detailed in the following.
\begin{itemize}
\item \textit{Lack of context}: Most quote collections \cite{newell2018quote,vaucher2021quotebank, goel2018proposing} lack context information and solely provide the quotes and their originators. To provide more context information in \kg{}, we extract quotes from Wikiquote -- a "free online compendium of sourced quotes from notable people
"\footnote{\url{https://en.wikiquote.org/wiki/Main_Page}}.
\item \textit{Tedious extraction process}: Even though Wikiquote is a semi-structured resource, extraction of quotes and contexts is a tedious process. In particular, we must design an extraction pipeline that is flexible across languages and adopts their characteristics. For example, it is necessary to differentiate the quotes not said by a person, but said about a person (e.g., English: "Quotes about Albert Einstein", German: "Zitate mit Bezug auf Albert Einstein").
\item \textit{Missing alignment of quote mentions:} As quote mentions in Wikiquote are not linked across languages, another important step is cross-lingual quote alignment which we perform using a language-agnostic transformer model that we evaluate on a ground truth set of manually aligned quote clusters.
\end{itemize}

Our contributions are as follows: (i) We propose a schema to represent quotes and context information. (ii) We propose an extraction pipeline that extracts quotes, their mentions and context information from all Wikiquote language versions. (iii) We align quote mentions across languages using a cross-lingual language model. (iv) We make \kg{} publicly available\footnote{\url{https://www.quotekg.l3s.uni-hannover.de}}.

The remainder of this paper is structured as follows: First, we describe the impact of \kg{} in the fields of Semantic Web, Natural Language Processing, Digital Humanities and others in more detail. Then, in Section~ \ref{sec:schema}, we describe the schema adopted for \kg{}. In Section \ref{sec:extraction}, we describe the \kg{} creation pipeline. In Section \ref{sec:examples}, we provide statistics and examples of \kg{}, followed by information about the availability and maintenance in Section~\ref{sec:availability}. Section \ref{sec:related_work} gives an overview of related work. Finally, we provide a conclusion in Section \ref{sec:conclusion}.

%% file: 02_impact.tex
\section{Potential Impact}
\label{sec:impact}

\kg{} contains quotes, a new type of information that is, to the best of our knowledge, not yet present in existing knowledge graphs. Therefore, \kg{} can potentially attract new audiences from several fields such as Digital Humanities and Natural Language Processing. While existing cross-domain or event-centric knowledge graphs such as Wikidata~\cite{vrandevcic2014wikidata}, DBpedia~\cite{auer2007dbpedia}, and EventKG~\cite{gottschalk2019eventkg} target the representation of real-world entities, including persons of public interest and important events, they scarcely represent what people actually said -- even though this information reflects persons' characteristics and can lead to an understanding of how particular events unfolded in the real world. Instead, facts about persons in knowledge graphs (e.g., properties representing birth dates, marriages, and awards received) without a doubt represent relevant facts in a person's life but typically do not reveal personal traits or surprising insights.
Existing corpora of quotes like Quotebank~\cite{vaucher2021quotebank} and the QUOTES500K dataset~\cite{goel2018proposing} provide large collections of English quotes. In contrast to these corpora, \kg{} is a knowledge graph and provides societal relevant quotes and contexts in $55$ languages, links them to other knowledge graphs, and aligns quote mentions across languages.

Potential applications of \kg{} are manifold: (i) First and foremost, \kg{} can add a new dimension to the exploration and investigation of the lives of public figures. While the creation of biography timelines from knowledge graphs has been studied in the past \cite{gottschalk2020eventkg,althoff2015timemachine}, such timelines do not consider the inclusion of quotes. \kg{} can help to enrich such timelines with relevant quotes to make them more lively and informative. (ii) Similarly, the analysis of quotes related to a specific topic over time can support research in the fields of Digital Humanities, and can be used to gauge public opinion regarding specific events. For example, there have been analyses of how social movements and global events affect language\cite{haun2020} and how the words used by public persons carry political backgrounds \cite{viala2020}. (iii) Quotes also play an important role when observing information propagation \cite{sims2020measuring} and the bias potentially caused by one-sided selection of quotes~\cite{niculae2015quotus}. (iv) \kg{} can also serve as an additional resource for machine translation, given that it contains $38,931$ quotes with mentions in different languages. (v) \kg{} can help answering questions such as "Who said 'yes, we can'?". (vi) \kg{} contains $13,104$ quotes labelled as misattributed or falsely claimed. They can be used as a resource for understanding the propagation of false or misleading information \cite{robinson2018did}. (vii) Finally, \kg{} can take quote collections to a new level: There are plenty of websites available that provide collections of quotes\footnote{\url{https://www.brainyquote.com}, \url{https://www.goodreads.com/quotes}, \url{https://www.successories.com/iquote}, \ldots}, typically monolingual and primarily for entertainment purposes (e.g., images of inspirational quotes or quote mashup games\footnote{\url{http://natetyler.github.io/}}). The context information in \kg{} can support the exploration and the search for quotes and provide important information surrounding a quote, and, as such, broaden the user's horizon.

%% file: 03_schema.tex
\section{\kg{} Schema}
\label{sec:schema}

The goal of the \kg{} schema is to model quotes, their relationships with persons and other entities, as well as their different mentions, e.g. translations, typically in different contexts. To this end, \kg{} is based on an extension of the \textit{schema.org} vocabulary that provides a \voc{so}{Quotation}\footnote{so: \url{https://schema.org/}} class which is re-used. According to the \url{schema.org} description, the \voc{so}{Quotation} class models quotes that are ``Often but not necessarily from some written work'' and can also refer to a ``Quotation from an Event''\footnote{\url{https://schema.org/Quotation}}. Therefore, it fits well to our concept of a quote in \kg{}. However, we extend the schema with a new class \voc{qkg}{Mention} which models the different mentions of a quote.

\begin{figure}[ht]
    \centering
    \includegraphics[width=\textwidth]{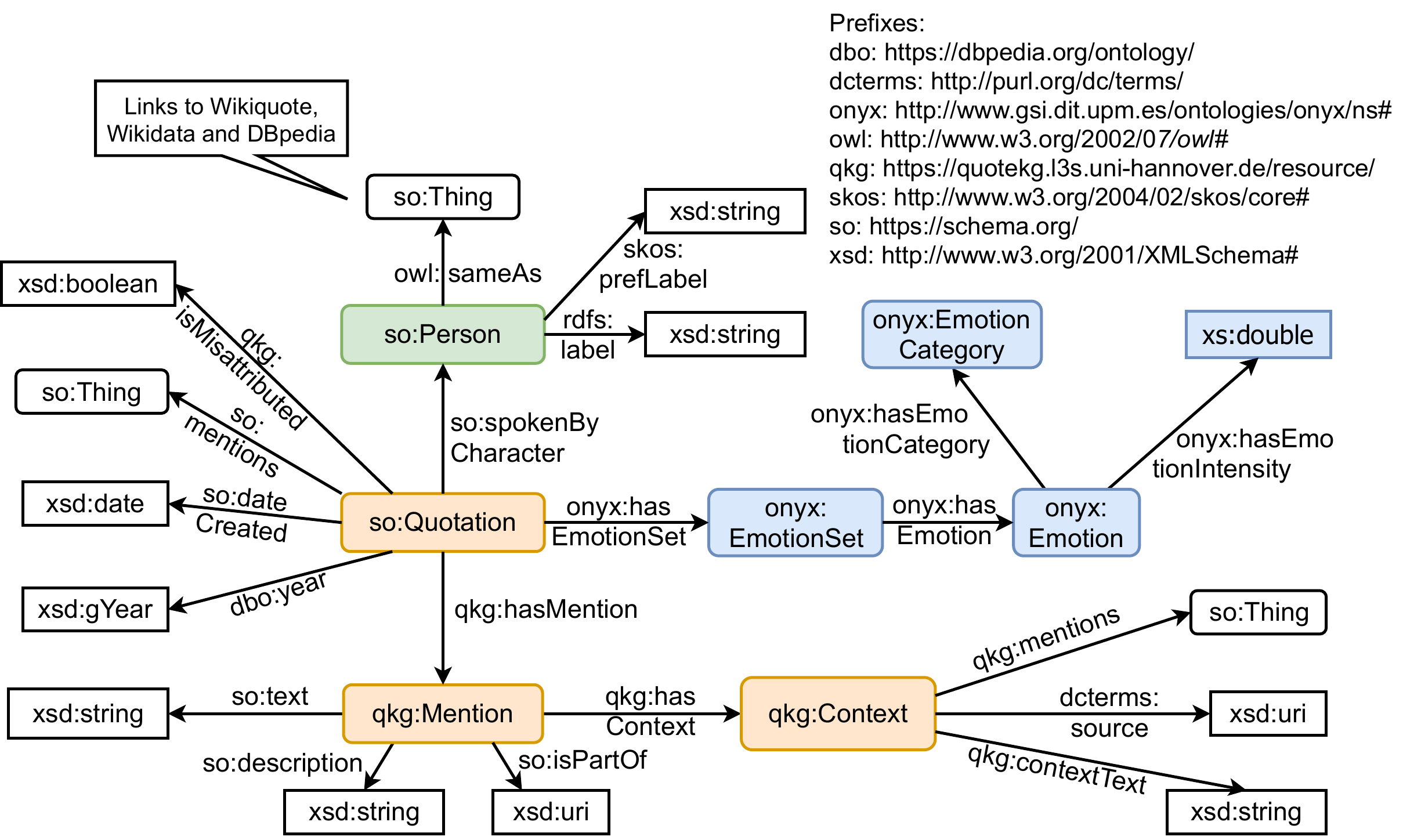}
    \caption{The \kg{} schema based on \textit{schema.org}. Arrows visualize the \voc{rdfs}{domain} and \voc{rdfs}{range} restrictions on properties. Name spaces and prefixes are described in the top right corner. Orange classes are related to quotes and their mentions, blue ones to the sentiment of a quote and the green class is about the person.}
    \label{fig:schema}
\end{figure}

Fig.~\ref{fig:schema} presents \kg{}'s schema. Its classes are described in the following.

\begin{itemize}
    \item \textbf{Person}: Each quote in \kg{} is assigned to a person modeled as \voc{so}{Person}. For persons, \kg{} provides additional type information (e.g., Politician) plus \voc{owl}{sameAs} relations to Wikidata and the different DBpedia and Wikiquote language editions.
    \item \textbf{Quote}: In \kg{}, a resource typed as \voc{so}{Quotation} refers to the unique event of something being said by a person of public interest (\voc{so}{spoken\-By\-Character}) at a specific point in time (\voc{so}{date\-Created}). A quote may also refer to other entities (\voc{so}{mentions}) of any type.
    \item \textbf{Mention}: A quote can be mentioned in different contexts: For example, there may be translations of the quote in different languages, alternative records of the same quote, or different contexts that a quote is extracted from. Therefore, we introduce the class \voc{qkg}{Mention}. Mentions can be related to one or more \voc{qkg}{Context} objects.
    \item \textbf{Context}: The context of a mention provides additional attributes that come together with the specific mention. For example, its origin (e.g., a reference to a specific interview) and the original source (e.g., a link to a news website). To model context, we create the class \voc{qkg}{Context}.
     \item \textbf{Sentiment}: For each quote, we provide its sentiment using the Onyx ontology which is used for describing emotions \cite{sanchez2016onyx}. A quote is assigned a score for a specific emotion category (``neutral'', ``negative'' or ``positive'').
\end{itemize}

\begin{figure}[t!]
    \centering
    \includegraphics[width=\textwidth]{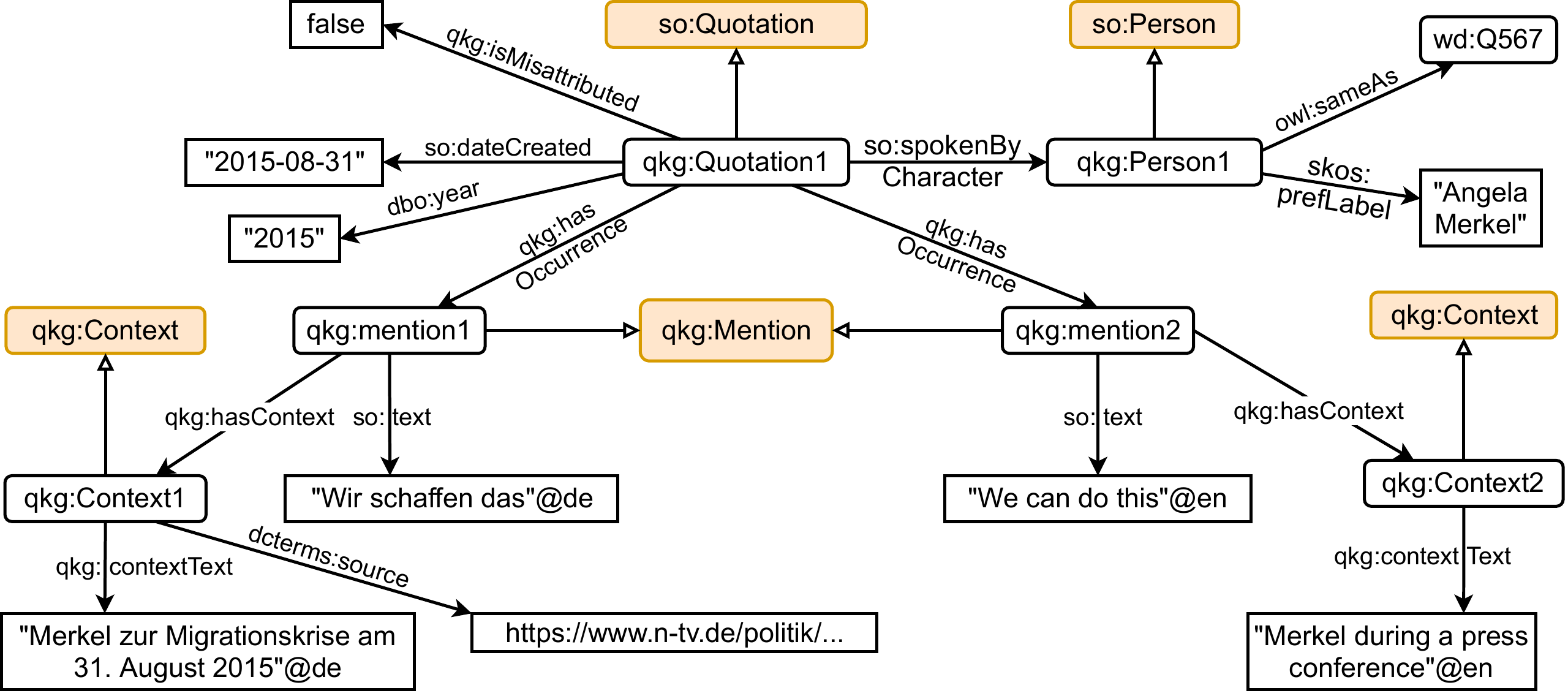}
    \caption{An example quote modeled using the \kg{} schema. $\rightarrowtriangle$ marks \voc{rdf}{type} relations. \voc{xsd} data type annotations were omitted for brevity. The prefixes and name spaces are the same as in Fig. \ref{fig:schema}, plus \schema{wd}: https://www.wikidata.org/entity/.}
    \label{fig:schema_example}
\end{figure}

Fig.~\ref{fig:schema_example} shows an example instantiation of the \kg{} schema. The quote "Wir schaffen das" introduced in Table~\ref{tab:intro_examples} is connected to two instances of \voc{qkg}{Mention}, one representing a German mention, the other one an English one ("We can do this"). Both mentions come with additional context information.

%% file: 04_extraction.tex
\section{Extraction and Alignment of Quotes}
\label{sec:extraction}

\begin{figure}[ht]
    \centering
    \includegraphics[width=\textwidth]{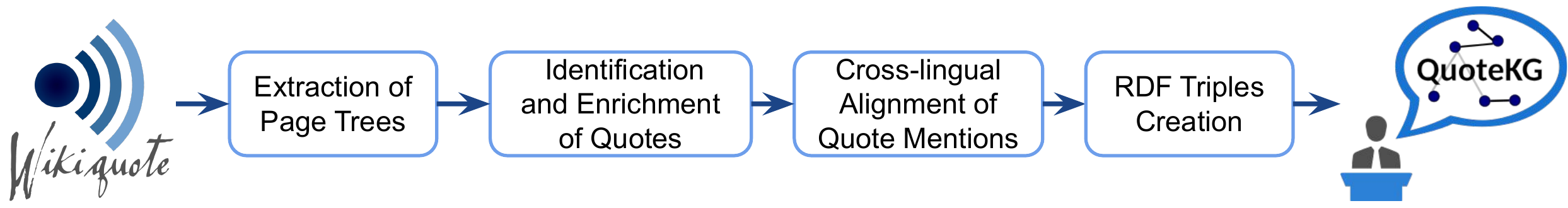}
    \caption{Pipeline to create \kg{} from Wikiquote.}
    \label{fig:pipeline}
\end{figure}

This section describes the input data and the implementation of the four main steps of the \kg{} creation pipeline shown in Fig.~\ref{fig:pipeline}.

\subsection{Wikiquote}
\label{subsec:wikiquote}

We base \kg{} on Wikiquote -- an online collection of quotes\footnote{\url{https://en.wikiquote.org/wiki/Main_Page}}. Wikiquote has a similar structure to Wikipedia: Independent versions of Wikiquote exist for different languages. Wikiquote contains pages, each of them about a given topic and divided into different sections and subsections. For \kg{}, we focus on Wikiquote pages about persons that contain quotes attributed to them. Example pages are the English\footnote{\url{https://en.wikiquote.org/wiki/Albert_Einstein}} and French page about Albert Einstein\footnote{\url{https://fr.wikiquote.org/wiki/Albert_Einstein}}.

Each Wikiquote page is formatted using the MediaWiki markup\footnote{\url{https://www.mediawiki.org/wiki/Help:Formatting}} and contains semi-structured content that includes the person's description, sections with quotes, references and more. The quotes are given in one of the following representations: in the traditional MediaWiki markup as shown in Fig.~\ref{fig:untemplated} or using pre-defined \textit{templates} that allow for a more structured definition of key-value pairs. For example, Fig.~\ref{fig:templated} shows the key-value pair (\textit{key}: \texttt{Citation}, \textit{value}: \texttt{Tomber amoureux\dots}).

\begin{figure}[t]
\fontsize{8pt}{9pt}\selectfont
    \centering
\begin{Verbatim}[frame=single]
* '''Falling in love is not at all the most stupid thing that people do
— but gravitation cannot be held responsible for it.'''
** Jotted (in German) on the margins of a letter to him (1933), p. 56
\end{Verbatim}
    \caption{Example of a quote in the English Wikiquote, based on MediaWiki markup.}
    \label{fig:untemplated}
\end{figure}

\begin{figure}[t]
\fontsize{8pt}{9pt}\selectfont
    \centering
\begin{Verbatim}[frame=single]
{{Citation|Tomber amoureux n'est pas du tout la chose la plus stupide
que font les gens — mais la gravitation ne peut en être tenue pour
responsable. |original=Falling in love is not at all the most stupid
thing that people do — but gravitation cannot be held responsible for
it.|langue=en}}
\end{Verbatim}
    \caption{Example of a quote in the French Wikiquote, using templates.}
    \label{fig:templated}
\end{figure}

While there are links between the pages describing persons in different languages, quote mentions are not linked across languages. Fig.~\ref{fig:untemplated} and Fig.~\ref{fig:templated} show two mentions of the same quote by Albert Einstein. The first is from the English Wikiquote and shows an English quote, the second one from the French Wikiquote is given in French and English. The original German quote is not available in these two language versions.

In general, one can observe a large imbalance in Wikiquote regarding the covered persons and the number of quotes in different language versions. This imbalance can often be explained by the different sizes of Wikiquote language versions and the difference in the cultural significance of a person in one language community compared to another. For example, there exists a French page with $35$ quotes and an Italian page with $2$ quotes of the former footballer Michel Platini who used to play in Italy, but there is no English page. This imbalance also implies that there is no guarantee that Wikiquote will contain the original language version of a quote. \kg{} can have multiple quote mentions of the same quote through cross-lingual quote mention alignment.

\subsection{Extraction of Page Trees}

\begin{figure}[t]
    \centering
    \includegraphics[width=\textwidth]{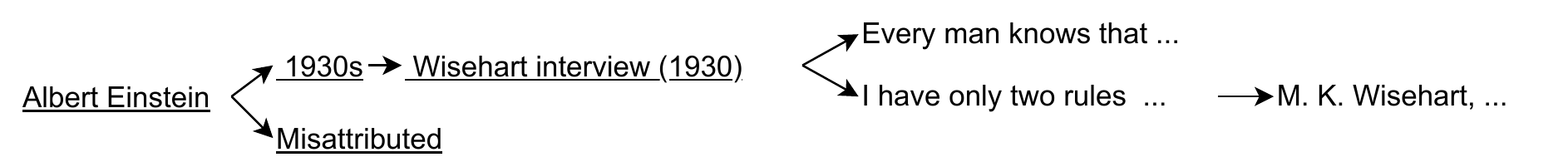}
    \caption{Excerpt of an example page tree from the English Wikiquote page about Albert Einstein. Section titles are underlined.}
    \label{fig:page_tree}
\end{figure}

In the beginning, our \kg{} creation pipeline processes all Wikiquote language editions with at least 50 pages, excluding Simple English\footnote{For more detailed statistics about Wikiquote language editions see: \url{https://wikistats.wmcloud.org/display.php?t=wq}} and selects all pages about persons.  
From each Wikiquote page about a person, we create a \textit{page tree}. The page tree consists of section titles plus quotes and contexts. An example page tree is presented in Fig.~\ref{fig:page_tree}.

\subsection{Identification and Enrichment of Quotes}

In the second step of the \kg{} creation pipeline, the page trees are transformed into a set of quotes with contextual information. To this end, we specify language-specific rules and enrich quotes and contexts with additional metadata.

To identify quotes, we first define a language-specific list of section titles denoting quotes (e.g., "Citations", "Zitate", "Citazioni") and contextual information (e.g., "útskýring", "Viitattu", "vydavatel"). In addition, we collect a list of template types representing quotes and consider all child nodes of section titles as quotes. From section titles and templates, we further gather the following:

\begin{itemize}
\item Dates: 
We identify the dates of quotes from a pre-defined list of template keys (e.g., ``année d'origine'' in French) for quotes extracted from templates. If such dates are not available or when dealing with quotes not extracted from templates, we extract dates from the section titles above the particular quote in the page tree and the contexts below the quote.

We select the time expression with the highest level of precision (e.g., we select May 2020 over 2020). In case of conflicts, no date is chosen.
\item Veracity: To reflect the authenticity of quotes and their contextual information, we capture whether a quote has been misattributed to the person. In Wikiquote, misattributed quotes are grouped into specified sections. We identify such sections with a manually created list of regular expressions (e.g., ``Misattributed'' (English) and ``Fälschlich zugeschrieben'' (German).
\item Sources: Often, context contains links to websites where the quote was reported. We collect such external links from templates and from the markup.
\item Linked entities: Quotes can be linked to entities such as other persons or organisations. We collect such links from templates and the markup.
\item Language: While the Wikiquote pages are written in specific languages, their quotes can be written in their original language or translated. For this reason, we use language detection to designate the language of a quote and do not rely on the language of the page tree.
\item Sentiment: We detect the sentiment of each quote mention (\textit{positive}, \textit{negative} or \textit{neutral} with a score between $0$ and $1$) using XLM-RoBERTa-Twitter, an XLM-RoBERTa model trained on $\sim198M$ multilingual tweets~\cite{barbieri2021xlm}.
\item Identity links: To establish \voc{owl}{sameAs} links between the QuoteKG entities, Wikidata and DBPedia, we use Wikidata's sitelinks\footnote{\url{https://www.wikidata.org/wiki/Help:Sitelinks/en-gb}}.
\end{itemize}

For all persons and entities identified during this process, we extract additional information regarding their labels and types from DBpedia and Wikidata.

\subsection{Cross-lingual Alignment of Quote Mentions}

After identifying and enriching quotes, we need to detect which of them represent mentions of the same quote said by a person of public interest. This task of cross-lingual alignment of quote mentions is treated as a clustering task at the end of which each cluster represents a quote with a set of mentions.

In detail, the clustering task is performed for each person in isolation. Given a person's quote mentions in a set of languages, we aim at creating clusters of highly similar mentions. 
To derive a similarity between two mentions, potentially from different languages, we compute the cosine similarity of sentence embeddings derived from the mentions' texts. As an embedding model, we use a language-agnostic transformer model pre-trained on millions of multilingual paraphrase examples in more than $30$ languages, namely XLM-RoBERTa \cite{conneau2019unsupervised}. The ability of such models to adapt to previously unknown languages has been shown in \cite{hu2020xtreme}. Given such embeddings and the cosine similarity function, clustering is performed by detecting communities of quotes using a nearest-neighbour search. To do so, we chose UKPLab's Fast Clustering algorithm\footnote{\url{https://github.com/UKPLab/sentence-transformers/blob/master/examples/applications/clustering/}} that is optimised towards efficient similarity computations of our embeddings.

To aggregate the sentiments of all mentions in a cluster, we take the most frequent sentiment category and average over the scores of that category.

\subsection{RDF Triples Creation}

After identification of quotes and their contexts and cross-lingual alignment, we transform them into RDF triples following the schema presented in Fig.~\ref{fig:schema}.

\subsection{Implementation}

We use the MWDumper\footnote{\url{https://www.mediawiki.org/wiki/Manual:MWDumper}} to process the Wikiquote XML dumps and parse the single pages given in the Wikipedia markup using the Bliki engine\footnote{\url{https://github.com/axkr/info.bliki.wikipedia_parser}}. For language detection and time expression extraction, we use the langdetect\footnote{\url{https://pypi.org/project/langdetect/}} and dateparser\footnote{\url{https://github.com/scrapinghub/dateparser/}} libraries. The Fast Clustering algorithm was run with a cosine similarity threshold of $0.8$. The creation of knowledge graph triples and their serialisation is done via the RDFLib library\footnote{\url{https://github.com/RDFLib/rdflib}}. The Java implementation of the dumper and the Python code for cross-lingual alignment and knowledge graph creation are publicly available on GitHub\footnote{\url{https://github.com/tkuculo/QuoteKG}}.

%% file: 05_evaluation_statistics_examples.tex
\section{Statistics, Evaluation, Examples \& Web Interface}
\label{sec:examples}

In this section, we first provide general statistics of \kg{}, evaluate the cross-lingual alignment and present example queries. 

\subsection{Statistics}

In total, \kg{} contains $880,878$ quotes with $961,535$ quote mentions. For $411,912$ mentions, context is available. Table~\ref{tab:statistics} provides detailed statistics for selected languages. \kg{} covers both high-resource languages such as English ($271,541$ quote mentions from $19,073$ persons) and Italian ($146,103$ quote mentions from $18,803$ persons), as well as low-resource languages such as Welsh ($508$ quote mentions from $239$ persons).

\begin{table}[t]
\footnotesize
\centering
\caption{Statistics of selected languages in \kg{}.}
\label{tab:statistics}
\begin{tabular}{lrrrr} \toprule
\multicolumn{1}{c}{\textbf{Language}} & \multicolumn{1}{c}{\textbf{Persons}} & \multicolumn{1}{c}{\textbf{Quotes}} & \multicolumn{1}{c}{\textbf{Mentions}} & \multicolumn{1}{c}{\textbf{Mentions with Contexts}}
 \\ \midrule
English & 19,073 & 267,740 & 271,541 & 193,848 \\
Italian & 18,803 & 145,235 & 146,103  & 48,107 \\
German & 3,461 & 16,012 & 16,441  & 4,330 \\
Croatian & 2,707 & 11,023 & 12,965 & 2,045 \\
Welsh & 239 & 461 & 508 & 247 \\ \midrule
\textbf{All Languages} & 69,467 & 880,878 & 961,535 & 411,912 \\ \bottomrule
\end{tabular}
\end{table}

\subsection{Evaluation of the Cross-lingual Alignment}

We evaluate the quality of the cross-lingual alignment of quote mentions by comparing to a ground truth of correctly clustered mentions. Creating such a ground truth is a tedious process due to the large amount of possible clusterings and the number of pairwise comparisons\footnote{When considering a person that has $10$ quotes in $5$ languages each, there are $\sum_{i}^{5-1} 10 \cdot i^2 = 1,000$ possible pairwise comparisons.}. We have selected eight persons with quotes in English, German and Italian and manually clustered their mentions. Ground truth clusters were then compared to the \kg{} clusters by viewing the clustering process as a series of decisions for each of the pairs of mentions~\cite{schutze2008introduction}. For example, we consider three positive pairs for a quote mentioned in three languages: (Mention$_1$, Mention$_2$), (Mention$_1$, Mention$_3$), (Mention$_2$, Mention$_3$).

Table~\ref{tab:evaluation} shows the results of this evaluation: Cross-language alignment in \kg{} shows an average precision of $1.0$ and an F$_1$ score of $0.99$ for this ground truth data set. Following the imbalance of Wikiquote's coverage described in Section~\ref{subsec:wikiquote}, there is a high number of true negatives, i.e., the majority of quotes are only mentioned once in all Wikiquote language versions. In total, there are only two mentions which are not clustered together but should have been. All the other clusters are correct.

\begin{table}[ht]
\centering
\footnotesize
\caption{Evaluation of cross-lingual alignment for eight selected persons in English, German and Italian. TP: true positives (mention pairs that were correctly clustered together) TN: true negatives (mention pairs that were correctly not clustered together), FP: false positives, FN: false negatives, P: precision, R: recall, F$_1$: F$_1$ score.}
\label{tab:evaluation}
\begin{tabular}{@{}lrrrrrrrr@{}}
\toprule
\textbf{Person} & \textbf{TP} & \textbf{TN} & \textbf{FP} & \textbf{FN} & \textbf{P} & \textbf{R} & \textbf{F$_1$} \\ \midrule
Alan Turing & 10 & 935 & 0 & 1 & 1 & 0.91 & 0.95 \\
Alexander the Great & 5 & 491 & 0 & 0 & 1 & 1.0 & 1.0 \\
Edward Snowden & 6 & 697 & 0 & 0 & 1 & 1.0 & 1.0 \\
Gustav Mahler & 1 & 44 & 0 & 0 & 1 & 1.0 & 1.0 \\
Jean-Claude Juncker & 4 & 776 & 0 & 0 & 1 & 1.0 & 1.0 \\
Marie Antoinette & 4 & 347 & 0 & 0 & 1 & 1.0 & 1.0 \\
Marie Curie & 2 & 251 & 0 & 0 & 1 & 1.0 & 1.0 \\
Tom Clancy & 1 & 2849 & 0 & 0 & 1 & 1.0 & 1.0 \\ \midrule
\textbf{Total} & 33 & 6390  & 0 & 1 & 1.0 & 0.99 & 0.99 \\ \bottomrule

\end{tabular}
\end{table}

Our ground truth set of manually aligned quote clusters is available on the \kg{} website.

\subsection{Example Queries}

In this section, we present two example queries demonstrating how to use \kg{} as a collection of quotes and as a resource to conduct research on the misattribution of quotes.

\subsubsection{\kg{} as a Collection of Quotes and their Originators} Listing~\ref{lst:sparql1} shows a SPARQL query that returns the five persons with the most quotes in\kg{}. Table~\ref{tab:example_query_1} shows these persons together with the number of quotes. Without surprise, the persons with most quotes are philosophers and writers, including Friedrich Nietzsche and Oscar Wilde, plus Albert Einstein, known for many (misattributed) quotes \cite{robinson2018did}.

\begin{lstlisting}[float=t,captionpos=b, caption=SPARQL query: Persons with the most quotes., label=lst:sparql1, frame=single]
SELECT ?Person (COUNT(?quote) AS ?NumberOfQuotes) WHERE {
  ?quote a so:Quotation ;
    so:spokenByCharacter [
       skos:prefLabel ?Person ] .
} GROUP BY ?Person
ORDER BY DESC(COUNT(?quote))
\end{lstlisting}

\begin{table}[ht]
\footnotesize
\centering
\caption{The first five results of the query in Listing~\ref{lst:sparql1}.}
\label{tab:example_query_1}
\begin{tabular}{@{}lr@{}}
\toprule
\multicolumn{1}{c}{\textbf{\texttt{?Person}}} & \multicolumn{1}{c}{\texttt{\textbf{?NumberOfQuotes}}} \\ \midrule
Friedrich Nietzsche & $2,530$ \\
Oscar Wilde & $1,786$ \\
Albert Einstein & $1,627$ \\
Donald Trump & $1,610$ \\
Johann Wolfgang von Goethe & $1,537$ \\ \bottomrule
\end{tabular}
\end{table}

\subsubsection{Verification of Quotes} Misinformation on the Internet has become an increasingly important problem and requires methods that classify the veracity of information \cite{thorne2018automated} and benefit from knowledge graphs such as ClaimsKG that provide annotated and erroneous facts \cite{tchechmedjiev2019claimskg}. While ClaimsKG provides wrong claims stated by persons extracted from fact-checking sites, \kg{} has quotes labelled as wrongly attributed to persons, thus a different type of misinformation.
The query shown in Listing~\ref{lst:sparql_fake} returns quotes of Albert Einstein that are marked as misattributed in \kg{} (see Table~\ref{tab:example_query_fake}), together with context information. Such context information can be a valuable resource for explaining misattribution in the case of quotes.

\begin{lstlisting}[float=t,captionpos=b, caption=SPARQL query: Quotes misattributed to Albert Einstein and their contexts., label=lst:sparql_fake, frame=single]
SELECT ?Text (SAMPLE(?contextText) AS ?contextTexts)
(SAMPLE(?source) AS ?Source) WHERE {
  ?quote so:spokenByCharacter [
    skos:prefLabel "Albert Einstein" ] ;
  qkg:isMisattributed true ; qkg:hasMention ?mention .
  
  ?mention so:text ?Text ;
    qkg:hasContext [
    qkg:contextText ?contextText ; so:source ?source
  ] .
} GROUP BY ?Text
\end{lstlisting}

\begin{table}[ht]
\footnotesize
\caption{Two results of the query in Listing~\ref{lst:sparql_fake}, returning quotes that were misattributed to Albert Einstein. Texts are shortened for brevity here.}
\label{tab:example_query_fake}
\begin{tabular}{@{}p{4.5cm}p{4cm}p{3cm}@{}}
\toprule
\multicolumn{1}{c}{\textbf{\texttt{?Text}}} & \multicolumn{1}{c}{\textbf{\texttt{?contextTexts}}} & \multicolumn{1}{c}{\textbf{\texttt{?Source}}} \\ \midrule
Everything is energy and that's all there is to it \dots It can be no other way. This is not philosophy. This is physics. & There's no evidence that Einstein ever said this. & http://quoteinvesti\-gator.com/2012/05/\-16/everything-energy/

\\
If the facts don't fit the theory, change the facts. & The earliest published attribution of this quote to Einstein found on \dots, but no source to Einstein's original writings is given \dots & http://\-books.\-google.\-com/\-books?id=... \\ \bottomrule
\end{tabular}
\end{table}

\subsection{Web Interface}

\begin{figure}[t]
    \centering
    \includegraphics[width=\textwidth]{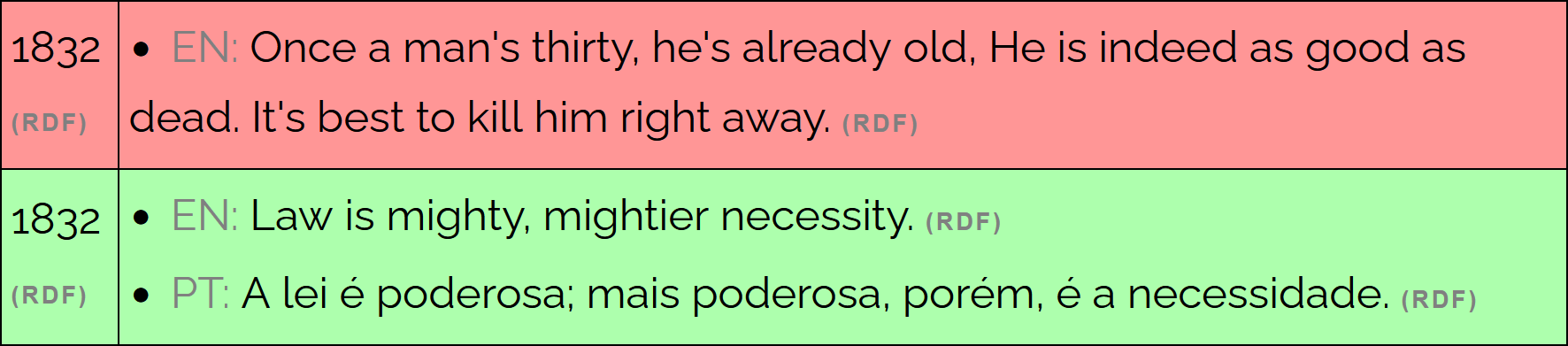}
    \caption{An example of the Search \& Demo interface showing two quotes of Johann Wolfgang Goethe which are available in Portuguese or English. The sentiment of quotes is indicated by colour (red: negative, green: positive).}
    \label{fig:interface}
\end{figure}

On the \kg{} website, we offer a SPARQL endpoint and a demo Search \& Demo interface where users can search for specific persons and display their quotes in selected languages. An example of this interface is shown in Fig.~\ref{fig:interface} which display Portuguese and English quotes of Johann Wolfgang von Goethe.

%% file: 07_availability.tex
\section{Availability}
\label{sec:availability}

\textbf{Availability:} The \kg{} website\footnote{\url{https://quotekg.l3s.uni-hannover.de}} provides access to a description of \kg{} and its schema, to the SPARQL endpoint, to data downloads and will provide a canonical citation to this paper. \kg{} is licensed under the Creative Commons Attribution Share Alike 4.0 International\footnote{\url{https://creativecommons.org/licenses/by-sa/4.0/legalcode}} license. Persistent access to the \kg{} triple files is provided through an upload to the zenodo repository\footnote{\url{https://zenodo.org/record/4702544}}.
The code for the creation of \kg{} is publicly available on GitHub\footnote{\url{https://github.com/tkuculo/QuoteKG}} and is licensed under the MIT license\footnote{\url{https://opensource.org/licenses/MIT}}.

\textbf{Sustainability plan:} To account for updates in all Wikiquote language editions, we plan to release new versions of \kg{} twice a year. To do so, we deploy a script that covers the entire pipeline depicted in Fig. \ref{fig:pipeline}, from the download of Wikiquote dumps in each language until the creation of triples.

\textbf{Adherence to standards:} \kg{} is modeled through the Resource Description Framework. Its schema is an extension of \textit{schema.org}. We provide a machine-readable description of \kg{} using the VoID vocabulary\footnote{\url{https://www.w3.org/TR/void/}}. \kg{} adheres to the Linked Data Principles: resources can be looked up through their URIs and they are interlinked with Wikidata and DBpedia.

%% file: 08_related_work.tex
\section{Related Work}
\label{sec:related_work}

In this section, we give an overview of other corpora and knowledge graphs containing quotes, other usages of Wikiquote and about cross-lingual alignment.

\subsubsection{Quote Corpora} Many collections of quotes have been created and maintained, mainly mono-lingual and without semantic annotations. Since the release of its first edition in 1941, The Oxford Dictionary of Quotations \cite{knowles2009oxford} aims at providing ``the wit and wisdom of past and present'' with a focus on the provenance of quotes. Provenance of quotes is also an indispensable criterion in the Book of Fake Quotes \cite{boller1989}. There are few machine-readable monolingual quote collections\footnote{\url{https://www.kaggle.com/akmittal/quotes-dataset}}$^,$\footnote{\url{https://github.com/JamesFT/Database-Quotes-JSON}} \cite{newell2018quote,goel2018proposing,vaucher2021quotebank}. These corpora are typically monolingual and extracted from news. Consequently, while they may have a large amount of quotes, they lack a mechanism to ensure societal relevance of quotes as in Wikiquote. As a knowledge graph, \kg{} enables easy access to quotes and rich metadata.

\subsubsection{Quotes in Knowledge Graphs} While DBQuote \cite{piao2015dbquote} allows user annotations of quotes extracted from Twitter and Wikiquote through an ontology, it only covers two languages (English and Korean) and has not been made available. 
To the best of our knowledge, \kg{} is the first publicly available knowledge graph of quotes. Consequently, quotes have only been insufficiently covered in the Semantic Web: for example, Wikidata~\cite{vrandevcic2014wikidata} contains less than $400$ instances of the class ``Phrase''\footnote{\url{https://www.wikidata.org/wiki/Q187931}} that are attributed to an author or creator -- most of them only consisting of few words (e.g., ``cogito ergo sum'' and ``covfefe'').
Event-centric knowledge graphs such as EventKG \cite{gottschalk2019eventkg} provide an understanding of the human history and world-shaking events. They do not include include quotes that complement the deeds of public figures. Many applications based on knowledge graphs (e.g., for exploring the lives of persons of public interest \cite{gottschalk2020eventkg, althoff2015timemachine}) could immediately profit from the inclusion of quotes.

\subsubsection{Wikiquote} Until now, Wikiquote has rarely been used as a research corpus, presumably due to the required but tedious extraction process. One example is the work by Buscaldi et al. who manually tagged quotes of the Italian Wikiquote as humorous or not, and used their annotated corpus for training models for humour recognition \cite{buscaldi2007some}. Giammona et al. analysed the spread of ancient quotes in today's Web through Wikiquote \cite{giammona2019dalla} and Wikiquote was used for training the chatbot Poetwannabe~\cite{chorowski2018talker}. With \kg{}, we foresee to ease the access to quotes for a wide range of research questions.

\subsubsection{Cross-lingual Alignment}
Several studies have shown that different languages share similar statistical properties that can be used to learn cross-lingual alignments between two languages, even without relying on any form of bilingual supervision~\cite{chung2018unsupervised}. 
While most works and datasets address bilingual alignment \cite{jing2019bipar, hieber2014wikiclir, GottschalkD17}, there are only few works on cross-lingual alignment \cite{liang2020xglue}. \kg{} focuses on the specific task of cross-lingual alignment of quote mentions.

%% file: 09_conclusion.tex
\section{Conclusion}
\label{sec:conclusion}

In this paper, we presented \kg{} -- a novel, multilingual knowledge graph of quotes. We have presented the \kg{} schema based on schema.org as well as a pipeline that extracts quotes from the Wikiquote corpus and aligns them across languages. \kg{} is publicly available and includes nearly one million quotes quotes in $55$ languages, said by nearly $69,000$ people of public interest.

%% file: 00_main.bbl
\begin{thebibliography}{10}
\providecommand{\url}[1]{\texttt{#1}}
\providecommand{\urlprefix}{URL }
\providecommand{\doi}[1]{https://doi.org/#1}

\bibitem{althoff2015timemachine}
Althoff, T., Dong, X.L., Murphy, K., Alai, S., Dang, V., Zhang, W.:
  {TimeMachine: Timeline Generation for Knowledge-Base Entities}. In:
  Proceedings of the 21th ACM SIGKDD International Conference on Knowledge
  Discovery and Data Mining. pp. 19--28 (2015)

\bibitem{auer2007dbpedia}
Auer, S., Bizer, C., Kobilarov, G., Lehmann, J., Cyganiak, R., Ives, Z.:
  {DBpedia: A Nucleus for a Web of Open Data}. In: The semantic web, pp.
  722--735. Springer (2007)

\bibitem{barbieri2021xlm}
Barbieri, F., Anke, L.E., Camacho-Collados, J.: {XLM-T: A Multilingual Language
  Model Toolkit for Twitter}. arXiv preprint arXiv:2104.12250  (2021)

\bibitem{boller1989}
Boller~Jr, P.F., George~Jr, O.J., et~al.: {They Never Said It: A Book of Fake
  Quotes, Misquotes, and Misleading Attributions: A Book of Fake Quotes,
  Misquotes, and Misleading Attributions}. Oxford University Press, USA (1989)

\bibitem{buscaldi2007some}
Buscaldi, D., Rosso, P.: {Some Experiments in Humour Recognition using the
  Italian Wikiquote Collection}. In: International Workshop on Fuzzy Logic and
  Applications. pp. 464--468. Springer (2007)

\bibitem{chorowski2018talker}
Chorowski, J., Lancucki, A., Malik, S., Pawlikowski, M., Rychlikowski, P.,
  Zykowski, P.: {A Talker Ensemble: The University of Wroclaw’s Entry to the
  NIPS 2017 Conversational Intelligence Challenge}. In: The NIPS'17
  Competition: Building Intelligent Systems, pp. 59--77. Springer (2018)

\bibitem{chung2018unsupervised}
Chung, Y.A., Weng, W.H., Tong, S., Glass, J.: {Unsupervised Cross-modal
  Alignment of Speech and Text Embedding Spaces}. arXiv preprint
  arXiv:1805.07467  (2018)

\bibitem{conneau2019unsupervised}
Conneau, A., Khandelwal, K., Goyal, N., Chaudhary, V., Wenzek, G., Guzm{\'a}n,
  F., Grave, E., Ott, M., Zettlemoyer, L., Stoyanov, V.: {Unsupervised
  Cross-lingual Representation Learning at Scale}. arXiv preprint
  arXiv:1911.02116  (2019)

\bibitem{giammona2019dalla}
Giammona, C., Yanes, E.S.: {From Print to Digital Texts, from Digital Texts to
  Print. Indirect Tradition of Latin Classics on the Web}. Storie e Linguaggi
  \textbf{5}(1) (2019)

\bibitem{goel2018proposing}
Goel, S., Madhok, R., Garg, S.: {Proposing Contextually Relevant Quotes for
  Images}. In: European Conference on Information Retrieval. pp. 591--597.
  Springer (2018)

\bibitem{GottschalkD17}
Gottschalk, S., Demidova, E.: {MultiWiki: Interlingual Text Passage Alignment
  in Wikipedia}. {ACM} Trans. Web  \textbf{11}(1),  6:1--6:30 (2017)

\bibitem{gottschalk2019eventkg}
Gottschalk, S., Demidova, E.: {EventKG--the Hub of Event Knowledge on the
  Web--and Biographical Timeline Generation}. Semantic Web  \textbf{10}(6),
  1039--1070 (2019)

\bibitem{gottschalk2020eventkg}
Gottschalk, S., Demidova, E.: {EventKG+BT: Generation of Interactive Biography
  Timelines from a Knowledge Graph}. In: European Semantic Web Conference. pp.
  91--97. Springer (2020)

\bibitem{haun2020}
Haun, M.: {How social movements and global events are changing language in
  2020} (2020), accessed: 2021-12-03

\bibitem{hieber2014wikiclir}
Hieber, F.: {WikiCLIR: a Cross-lingual Retrieval Dataset from Wikipedia}.
  Universit{\"a}t (2014)

\bibitem{hu2020xtreme}
Hu, J., Ruder, S., Siddhant, A., Neubig, G., Firat, O., Johnson, M.: Xtreme: A
  massively multilingual multi-task benchmark for evaluating cross-lingual
  generalisation. In: International Conference on Machine Learning. pp.
  4411--4421. PMLR (2020)

\bibitem{jing2019bipar}
Jing, Y., Xiong, D., Zhen, Y.: {BiPaR: A Bilingual Parallel Dataset for
  Multilingual and Cross-lingual Reading Comprehension on Novels}. arXiv
  preprint arXiv:1910.05040  (2019)

\bibitem{keyes2007quote}
Keyes, R.: {The Quote Verifier: Who said What, Where, and When}. St. Martin's
  Griffin (2007)

\bibitem{Thoughts}
Khurana, S.: {These 4 Quotes Completely Changed the History of the World }.
  \url{https://www.thoughtco.com/quotes-that-changed-history-of-world-2831970},
  accessed: 2021-12-01

\bibitem{knowles2009oxford}
Knowles, E.: {The Oxford Dictionary of Quotations}. Oxford University Press
  (2009)

\bibitem{wirschaffendas}
Krämer, A.: {Ein Satz mit Folgen}.
  \url{https://www.tagesschau.de/inland/merkel-wir-schaffen-das-109.html}
  (2021), accessed: 2021-12-01

\bibitem{liang2020xglue}
Liang, Y., Duan, N., Gong, Y., Wu, N., Guo, F., Qi, W., Gong, M., Shou, L.,
  Jiang, D., Cao, G., et~al.: {XGLUE: A New Benchmark Dataset for Cross-lingual
  Pre-training, Understanding and Generation}. arXiv preprint arXiv:2004.01401
  (2020)

\bibitem{mushaben2017wir}
Mushaben, J.M.: {Wir schaffen das! Angela Merkel and the European refugee
  crisis}. German Politics  \textbf{26}(4),  516--533 (2017)

\bibitem{newell2018quote}
Newell, C., Cowlishaw, T., Man, D.: Quote extraction and analysis for news. In:
  Proceedings of the Workshop on Data Science, Journalism and Media, KDD.
  pp.~1--6 (2018)

\bibitem{niculae2015quotus}
Niculae, V., Suen, C., Zhang, J., Danescu-Niculescu-Mizil, C., Leskovec, J.:
  {Quotus: The Structure of Political Media Coverage as Revealed by Quoting
  Patterns}. In: Proceedings of the 24th International Conference on World Wide
  Web. pp. 798--808 (2015)

\bibitem{piao2015dbquote}
Piao, G., Breslin, J.G.: {DBQuote: A Social Web based System for Collecting and
  Sharing Wisdom Quotes}. In: Proceedings of the 5th Joint International
  Semantic Technology Conference, Poster and Demonstrations (2015)

\bibitem{reucher2021}
Reucher, G.: {Famos Quotes: Why are so many fake?} (2021), accessed: 2021-12-03

\bibitem{robinson2018did}
Robinson, A.: {Did Einstein Really Say that?} Nature  \textbf{557}(7703),
  30--31 (2018)

\bibitem{sanchez2016onyx}
S{\'a}nchez-Rada, J.F., Iglesias, C.A.: {Onyx: A Linked Data Approach to
  Emotion Representation}. Information Processing \& Management
  \textbf{52}(1),  99--114 (2016)

\bibitem{schutze2008introduction}
Sch{\"u}tze, H., Manning, C.D., Raghavan, P.: {Introduction to Information
  Retrieval}, vol.~39. Cambridge University Press Cambridge (2008)

\bibitem{sims2020measuring}
Sims, M., Bamman, D.: {Measuring Information Propagation in Literary Social
  Networks}. In: Proceedings of the 2020 Conference on Empirical Methods in
  Natural Language Processing (EMNLP) (2020)

\bibitem{tchechmedjiev2019claimskg}
Tchechmedjiev, A., Fafalios, P., Boland, K., Gasquet, M., Zloch, M., Zapilko,
  B., Dietze, S., Todorov, K.: {ClaimsKG: A Knowledge Graph of Fact-checked
  Claims}. In: International Semantic Web Conference. pp. 309--324. Springer
  (2019)

\bibitem{thorne2018automated}
Thorne, J., Vlachos, A.: {Automated Fact Checking: Task Formulations, Methods
  and Future Directions}. In: Proceedings of the 27th International Conference
  on Computational Linguistics. pp. 3346--3359 (2018)

\bibitem{vaucher2021quotebank}
Vaucher, T., Spitz, A., Catasta, M., West, R.: {Quotebank: A Corpus of
  Quotations from a Decade of News}. In: Proceedings of the 14th ACM
  International Conference on Web Search and Data Mining. pp. 328--336 (2021)

\bibitem{viala2020}
Viala-Gaudefroy, J., Lindaman, D.: {Donald Trump’s ‘Chinese virus’: the
  politics of naming} (2020), accessed: 2021-12-03

\bibitem{vrandevcic2014wikidata}
Vrande{\v{c}}i{\'c}, D., Kr{\"o}tzsch, M.: {Wikidata: a Free Collaborative
  Knowledge Base}. Communications of the ACM  \textbf{57}(10),  78--85 (2014)

\end{thebibliography}
